\documentclass[letterpaper,twocolumn,10pt]{article}
\usepackage{usenix2019_v3}
\usepackage{graphicx}
\usepackage{wrapfig}

\hypersetup{%
  colorlinks=true,
  linkcolor=blue,
  linkbordercolor=red,
  citecolor=blue,
  linkcolor = blue,
  anchorcolor = blue,
  filecolor = blue,
  urlcolor = red
}

\begin{document}

\date{}

\title{\Large \bf An End-to-End ML System for Personalized Conversational Voice Models in Walmart E-Commerce}

\author{
{\rm Rahul Radhakrishnan Iyer\thanks{Corresponding Author: \texttt{rahul.iyer@walmartlabs.com}}}\\
Walmart Labs\\
Sunnyvale, CA
\and
{\rm Praveenkumar Kanumala}\\
Walmart Labs\\
Sunnyvale, CA
\and
{\rm Stephen Guo}\\
Walmart Labs\\
Sunnyvale, CA
\and
{\rm Kannan Achan   }\\
Walmart Labs\\
Sunnyvale, CA
} 

\maketitle

\begin{abstract}
Searching for and making decisions about products is becoming increasingly easier in the e-commerce space, thanks to the evolution of recommender systems. Personalization and recommender systems have gone hand-in-hand to help customers fulfill their shopping needs and improve their experiences in the process. With the growing adoption of conversational platforms for shopping, it has become important to build personalized models at scale to handle the large influx of data and perform inference in real-time. In this work, we present an end-to-end machine learning system for personalized conversational voice commerce. We include components for implicit feedback to the model, model training, evaluation on update, and a real-time inference engine. Our system personalizes voice shopping for Walmart Grocery customers and is currently available via Google Assistant, Siri and Google Home devices.
\end{abstract}

\section{Introduction}

There has been a growing need for shopping ease, especially in the e-commerce domain. With the advent of virtual assistants in the marketplace, users have started taking different avenues to obtain their everyday needs. It has been estimated that over 16\% of smart speaker owners shop monthly by voice and over 21\% of the consumers have engaged in voice shopping. With increasing adoption of conversational platforms, it has become paramount to personalize and cater the shopping experiences to the users' needs.

Conversational search and recommendation are relatively new research topics, but the basic concepts date back to some of the most early works in the community. For example, Croft and Thompson \cite{croft1987i3r} designed $I^{3}R$ (Intelligent Intermediary for Information Retrieval) – an expert intermediary system that takes activities to communicate with the user during a search session similar to what is done by a human intermediary; Belkin et al \cite{belkin1995cases} designed the MERIT system – an interactive information retrieval system that used script-based conversational interaction for effective search. Radlinski and Craswell \cite{radlinski2017theoretical} proposed a theoretical framework for conversational search, which described some basic design philosophies for conversational search systems. Kenter and de Rijke \cite{kenter2017attentive} formalized conversational search as a machine reading task for question answering. 

Yang et al \cite{yang2018response,yang2017neural} conducted next question prediction and response ranking in conversations. Spina and Trippas et al studied the ways of presenting search results over speech-only channels \cite{spina2017extracting} and transcribing the spoken search recordings \cite{trippas2017conversational} to support conversational search. Christakopoulou et al \cite{christakopoulou2016towards} proposed an interactive recommendation protocol that collects like/dislike feedback from users to refine the recommendations. Recently, several approaches involving natural language processing \cite{iyer2019event,iyer2019unsupervised,iyer2019heterogeneous,iyer2017detecting,iyer2019machine,iyer2017recomob,iyer2019simultaneous,iyer2020modeling,iyer2020transition}, machine learning \cite{li2016joint,iyer2016content,honke2018photorealistic,iyer2020correspondence,li2016joint2}, deep learning \cite{iyer2018transparency,li2018object} and numerical optimizations \cite{radhakrishnan2016multiple,iyer2012optimal,qian2014parallel,gupta2016analysis,radhakrishnan2018new} have also been used in the visual and language domains.

In this work, we present an end-to-end machine learning system for conversational voice models, particularly addressing conversational search, with a capability of real-time update. We have currently deployed several models for different conversational tasks using common feature sets: however, we focus on the specific use-case of search ranking. Our approach differs from existing prior works in the following ways
\begin{itemize}
    \itemsep-0.4em 
    \item We propose an end-to-end machine learning system for conversational voice models, including a seamless deployment system and real-time update of the models
    \item We include an implicit feedback system to personalize the voice shopping experience
    \item Our system has the ability to perform real-time inference with reduced latency
    \item We use a common feature store for different models deployed through the system
    \item Testing on held-out data before updating and pushing model to model store
\end{itemize}

\begin{figure}[!h]
  \centering
    \includegraphics[width=\columnwidth]{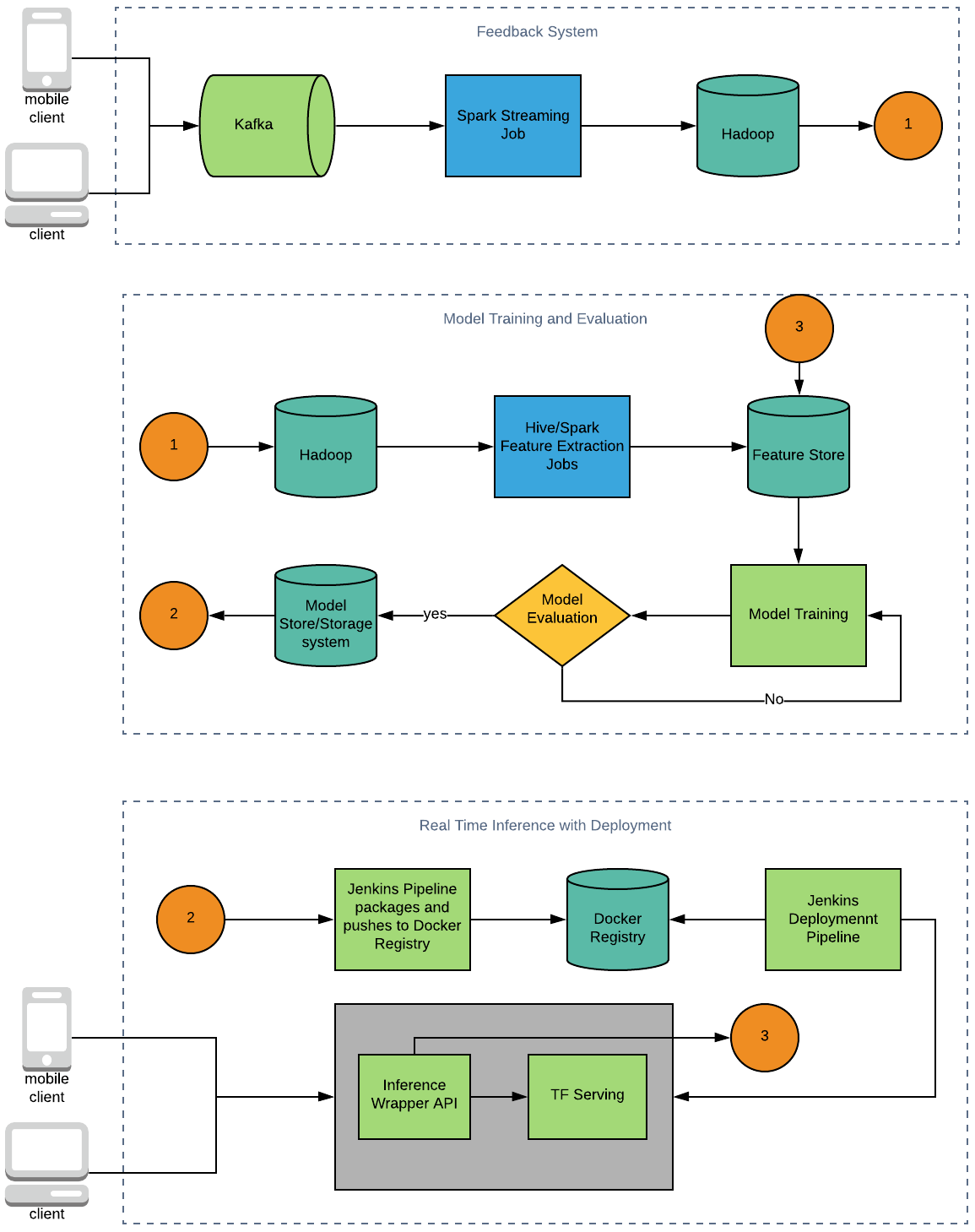}
  \caption{End-to-End ML System}
  \label{fig:deploy}
\end{figure}

\section{Overview of End-to-End ML System}
In this section, we provide a high level overview of the End-to-End ML system as illustrated in Figure \ref{fig:deploy}. This system is comprised of 3 main components and each individual component is  described below.

\subsection{Model Training and Evaluation}
We build feature sets from different sources using Hive/Spark Jobs. This is then stored in a distributed data store, which is used for both offline model training and Real-Time online (model) inference. The trained model is evaluated on a held-out standard dataset and pushed into the Model Store if performance is sufficient on evaluation metrics.

\subsection{Real-time Inference with Deployment}
This component consists of a set of pipelines, which are responsible for pulling the latest model from the model store, packaging into a docker image, and then pushing the custom image into the internal Docker Registry. The responsibility of the other pipeline is to take the latest image and deploy into the Real-Time inference system. The Real-Time inference system consists of an inference wrapper API, such as Flask, and TensorFlow Serving. The inference wrapper API connects to the feature store for a real-time fetch and call the TensorFlow Serving API with these features.

\subsection{Feedback System}
The Feedback system collects the click stream data and ATC data from clients, which is then stored on Hadoop. This feedback data is used in model training and updates.

\section{Use Case: Personalized Search Re-Ranking}
In this section, we describe how we utilize our system for the task of personalizing conversational voice search. The subsequent sections detail the feature sets used, modules in the architecture and model updates.

\subsection{Feature Sets}
We extract different feature sets from the users' interaction behaviors and item catalog information. We utilize a cache system to store these features to use across the different models:
\begin{itemize}
    \itemsep-0.4em 
    \item User Transactional, Interaction and Household Data
    \item Parsed Textual Search Queries with attributes and facets
    \item Product attributes and facets
\end{itemize}

These feature sets are built from the different sources using Hive and Spark jobs.

\subsection{Architecture}
Our voice search personalization architecture consists of the following five components: 
\begin{enumerate}
    \itemsep-0.4em 
    \item Data store (customer, product, query)
    \item Candidate product retrieval model
    \item Realtime Query parser
    \item Product attribute and facet extractor
    \item Tensorflow reranking model
\end{enumerate}

Real-time customer engagement signals and historical transactional data are used to build a preference signal for the user, which enables us to personalize for the voice channel. 

The following data is stored in cache: 1) customer transactional data, 2) parsed query attributes for the top searched textual queries, 3) parsed product attributes and facets for the catalog. We utilize the feature sets stored in cache to perform the personalized ranking. The tensorflow ranker performs a similarity computation between the search query and a list of items based on the attributes/facets, and ranks them. 

As soon as an update to the feature store is made, we retrain our model. It is then pushed to the model store using the pipelines described earlier if it meets the evaluation criteria.

\subsection{Feedback Integration}
We include a feedback loop in the system to capture explicit user preferences. If a suggestion is not accepted, the interaction behavior is used to modify the experiences accordingly.
\section{Merits of the System}
There are several merits to the system that we have deployed in production
\begin{itemize}
    \itemsep-0.4em
    \item Seamlessly evaluate and update models including an implicit feedback loop
    \item Ability to perform real-time inference with reduced latency
    \item Ability to use a common feature store for different models
\end{itemize}

\bibliographystyle{plain}
\bibliography{refs}

\end{document}